\documentclass[12pt,letterpaper]{article}

\usepackage[legalpaper, portrait, margin=1in]{geometry}
\usepackage{newtxtext}
\usepackage{newtxmath}
\usepackage{gensymb}
\usepackage{textcomp}
\usepackage[shortlabels]{enumitem}

\usepackage[pdftex]{graphicx}
\graphicspath{{./figures/}}
\DeclareGraphicsExtensions{.png}

\usepackage[dvipsnames]{xcolor}
\usepackage{amsmath}
\usepackage{algorithmic}
\usepackage{array}
\usepackage{threeparttable}
\usepackage{booktabs}
\usepackage{url}
\usepackage{cclicenses}
\usepackage{wrapfig}
\usepackage[utf8]{inputenc}
\usepackage{fontawesome5}
\usepackage[colorlinks,allcolors=black,pdftex]{hyperref}

\usepackage{biblatex} 
\begin{document}

\title{\textbf{Machine Learning Computer Vision Applications\\for Spatial AI Object Recognition\\in Orange County, California}}

\author{Dr.~Kostas~Alexandridis,~GISP\thanks{\textbf{~Kostas Alexandridis, PhD, GISP} is a Spatial Complex System Scientist/ GIS Analyst with the County of Orange, Orange County Public Works, OC Survey, Geospatial Services. Address: 601 W. Ross St, PO Box 4048, Santa Ana, CA 92701, USA, (email: \href{mailto:kostas.alexandridis@ocpw.ocgov.com}{Kostas.Alexandridis@ocpw.ocgov.com}). See \href{https://www.mindscribble.net/biography}{\textcolor{OliveGreen}{\faIcon{brain}~mindscribble.net}} or \href{https://www.linkedin.com/in/ktalexan/}{\textcolor{NavyBlue}{\faIcon{linkedin}~Kostas Alexandridis}} for more author information. Manuscript Created: April, 2021; Updated: January, 2023. This work was supported entirely from funding of the Orange County Public Works, OC Survey. This work is licensed under a \href{https://creativecommons.org/licenses/by-sa/4.0/legalcode.txt}{\textcolor{black}{\faIcon{creative-commons}~Creative Commons "Attribution-ShareAlike 4.0"}} license.
}}
\date{}
\maketitle

\begin{abstract}
We provide an integrated and systematic automation approach to spatial object recognition and positional detection using AI machine learning and computer vision algorithms for Orange County, California. We describe a comprehensive methodology for multi-sensor, high-resolution field data acquisition, along with post-field processing and pre-analysis processing tasks. We developed a series of algorithmic formulations and workflows that integrate convolutional deep neural network learning with detected object positioning estimation in 360° equirectancular photosphere imagery. We provide examples of application processing more than 800 thousand cardinal directions in photosphere images across two areas in Orange County, and present detection results for stop-sign and fire hydrant object recognition. We discuss the efficiency and effectiveness of our approach, along with broader inferences related to the performance and implications of this approach for future technological innovations, including automation of spatial data and public asset inventories, and near real-time AI field data systems.    
\end{abstract}

\section{Introduction}\label{sec1}

The twentieth-first century has introduced information-related functions in our everyday lives and social practices. For many of the youngest members of our societies throughout the world, life without a constant and readily accessible digital information flow is almost unimaginable. Of course, the latter is more profound in societies with higher degree of technological advancement. Nevertheless, information flows become more and more a core part of our contemporary realities. It has moved beyond the intellectual symbolism and realm of our social construction of reality, to become a well embedded norm and way of functioning in our societies. Information flows are in the core of our physical realities, infrastructure and \textit{modus operandi} of the functioning and organizing our social systems. Education, technology, science, governance and institutions across all aspects and facets of our social lives are dependent and rely upon digital information systems and flows.

The growing and relatively widespread use of artificial intelligence approaches to machine learning, has begun to proliferate applied and practical technological innovation across academic, government, industry, and commercial sectors alike. Technological advances in the last decade have completely transformed and restructure the dynamics and nature of technological innovation. The explosive growth of cloud technologies, methods, physical infrastructure, and software/algorithmic methods has exponentially reached a point where AI is becoming the industry’s gold standard from smart technologies. From smart homes and devices, smart vehicles, smart wearable devices, self-driving technologies and software, intelligent vehicle collision avoidance systems, industrial robotics, smart detection technologies, are just some examples of technological innovation with which large part of our societies are steadily growing accustomed with. Among these innovations, one area stands in the intersection of AI, machine learning and pattern recognition: deep learning methodologies, and its relevant applications for machine learning such as computer and cognitive vision. Such methods gained particular attention and attracted scientific and practical focus \cite{goodfellow_deep_2016} in the past few years.

This paper will introduce, describe, and discuss the application of machine learning technologies in acquiring, processing, and analyzing multi-sensor field imagery data for spatial AI object recognition tasks. Specifically, we present (a) a systematic methodology for field acquisition of multi-sensor imagery data (including integrated  \textsl{LiDAR} point cloud data, high-accuracy \textsl{GNSS} positional data and 360° photosphere imagery data); (b) a workflow frame-work for post-field and pre-analysis processing data tasks; (c) a machine learning computer vision data analysis framework for spatial object recognition and positional detection, and; (d) an automated workflow for GIS analysis and visualization of spatial data. We will provide examples of data analyzing though these integrated workflows and discuss our findings along with directions for next steps of analysis.

\section{Background}\label{sec2}

Deep Learning or, more descriptively, \textit{deep neural networks (DNN)} methodologies use auto-differentiated back propagation techniques to train a neural network classifier (using training and validation datasets consequently). A growing number of analytical and visualization methods for deep neural network learning have been developed in the last 5-10 years \cite{abiodun_state---art_2018,montavon_methods_2018,indolia_conceptual_2018,lecun_deep_2015}. An example of recent taxonomical classification of these methods is provided by Yu and Shi in their recent paper \cite{yu_user-based_2018}. For example, deep learning algorithms related to locational and/or spatially explicit situations have been designed and implemented, as in an application of DNN, where Yan et al \cite{yan_deep_2018} enabled the ability to predict vehicle speeds using data screening of historical velocity, acceleration, steering signal input, location and temporal awareness data from electric vehicle’s onboard GPS sensors. Convolutional neural network approaches as part of a deep learning framework have been reported in applications of 3D audio-visual patter recognition \cite{torfi_3d_2017}, scene pattern classification tasks in imagery \cite{weng_automatic_2019}, in satellite imagery analysis tasks using transfer learning \cite{wurm_semantic_2019}, color classification imagery tasks \cite{yin_quaternion_2019}, to name a few applications.

A few methodologies have been proposed, used and implemented in situations where traffic sign recognition is sought and pursued as part of an unsupervised or semi-supervised classification framework \cite{liu_machine_2019, arcos-garcia_evaluation_2018}. For example, Aziz et al \cite{aziz_traffic_2018} recently used an algorithmic methodology called extreme learning machine (ELM) to evaluate traffic sign classification performance in German (GTSRB) and Belgian (BTSC) sign datasets. The method uses gray-scale small resolution images with normalized band histogram intensity values to perform feature fusion for their ELM classifier. Their classification accuracy ranged between 95\% to more than 99\% (in the case of multi-dataset training sets). Yu et al \cite{yu_bag--visual-phrases_2016} used hierarchical deep models for traffic sign recognition in  \textsl{LiDAR} data analysis tasks with classification accuracy ranging from 85\%-99\% for different algorithmic implementations and datasets. Zhu et al \cite{zhu_traffic-sign_2016} performed traffic sign recognition tasks in panoramic images using convolutional neural networks for the German (GTSDB) benchmark database. Panorama images was used in Fakour-Sevom et al \cite{fakour-sevom_360_2018} and Coors et al \cite{ferrari_spherenet_2018}. Traffic sign detection and classification studies have been reported in a number of studies with varying results, including AdaBoost wavelet detection \cite{bahlmann_system_2005}, rectangle detection \cite{keller_real-time_2008},  \textsl{LiDAR} and image combinations \cite{guan_convolutional_2019, zhang_automated_2019, gong_frustum-based_2020}.

\section{Materials and Methods}\label{sec3}

Most of the studies and related literature described in the previous section uses benchmark pre-compiled databases and training datasets for developing and testing algorithmic implementations of deep learning and convolutional neural network models. Furthermore, most studies only address issues of detection, recognition, and classification rather than incorporating comprehensive spatial and locational integration of data. In many cases benchmark training datasets may incorporate data that are unsuitable for every situation, such as traffic signs from different countries, with different shapes and designs, or data with relative low resolutions that may not be suitable for high-resolution, high accuracy needs and settings. For example, high-accuracy \textsl{GNSS} systems require and mandate the use of high-resolution imagery for centimeter positional accuracy. We developed a comprehensive methodology that incorporates high resolution field data acquisition and collection, multi-task pre-processing and analysis, algorithmic detection, recognition, and classification tasks, and spatial and GIS locational processing and geodatabase construction of object detections. We describe the methodological components used in this study in the next subsections.

\subsection{LiDAR, GNSS and 360\textdegree~Image Photosphere Data Acquisition}\label{sec3.1}

\begin{figure}[ht]
    \centering
    \includegraphics[width=0.75\textwidth]{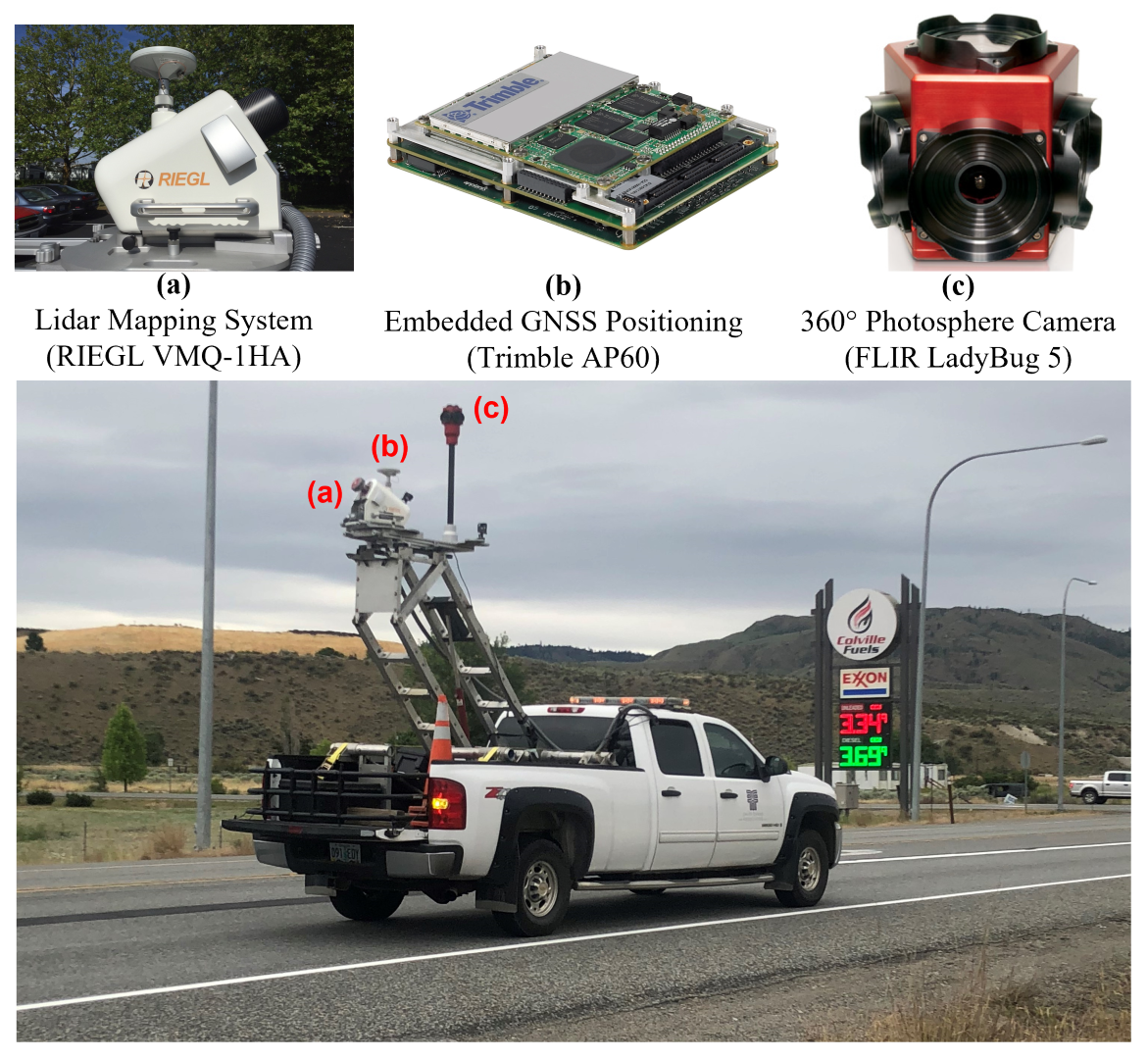}
    \caption{An overview of the field collection instruments used for the data collection in the study, that include a mobile  \textsl{LiDAR} sensor, a Trimble \textsl{GNSS} positioning system, and a photosphere image camera (credit: Matthew Kumpula, David Evans and Associates Inc).}
    \label{fig01}
\end{figure}

The field data collection was conducted by David Evans and Associates \cite{david_evans_and_associates_inc_surveying_2020} using the mobile unit configuration described in the next subsections (see also \figurename{~\ref{fig01}}). The field unit configuration included three integrated sensor systems: firstly, a mobile  \textsl{LiDAR} mapping system for capturing 3D point cloud data, as shown in \figurename{~\ref{fig01}(a)}; secondly, an embedded \textsl{GNSSS}-inertial positioning system for accurate locational data acquisition, shown in \figurename{~ref{fig01}(b)}, and; thirdly, an automated 360° photosphere image capture system, shown in \figurename{~\ref{fig01}(c)}.

The mobile  \textsl{LiDAR} data were acquired by a sensor amounted on a surveying vehicle and the point cloud imagery was captured as the vehicle traveled along the streets. The mobile unit used was the RIEGL VMQ-1HA high-speed single scanner mapping system \cite{riegl_laser_management_systems_gmbh_riegl_2019}. The tiled point clouds cover the scene of the whole 3D space along the streets, including the trees, grass, buildings, cars, pedestrian and so on. There are some points with z coordinates of negative values that corresponding to catch basins at both sides of the roads.

The \textsl{GNSS} positioning system used was an Trimble Applanix AP60 embedded on-board card along with an inertial measurement unit \cite{trimble_applanix_trimble_2016} for capturing high-accuracy GPS positioning data combining the vehicle’s real-time position and the photosphere imaging data.

The camera used for static equirectangular 360\textdegree~imagery acquisition was a FLIR Ladybug 5, 30MP (5MP\texttimes6 sensors) camera imaging system \cite{flir_integrated_imaging_solutions_flir_2017}. It produces full bit depth of 12-bit RAW images, converted to JPEG post-field collection using the camera’s API software in C\#. The camera’s field of view is 90\% of full sphere, and the spherical distance is calibrated at minimum focus distance of 2 meters (6.562 feet). The imaging system itself reports data from several on-board environmental sensors, namely temperature, barometer, humidity acceleration and compass readings for each capture.

Data collection was conducted by David Evans and Associates Inc \cite{david_evans_and_associates_inc_surveying_2020}. The data collection dates for the data reported in this paper ranged from December 2018, to April 2019 in 12 data sample groups and 8 field sample days. They cover two regions in Orange County: Anaheim Hills (3 field sample days in March and April 2018) and North Tustin (5 field sample days in March and April 2019), both unincorporated areas of the County of Orange (see results section \ref{sec4} for more details).

\subsection{Field Post-Processing Data Applications}\label{sec3.2}

The mobile  \textsl{LiDAR} point-cloud data underwent field post-processing using the MLS Ri Software suite from RIEGL (RiWORLD for georeferencing mobile data, RiPROCESS for data processing, RiANALYZE for wave-form analysis and RiPRECISION for mobile data adjustment) \cite{riegl_laser_management_systems_gmbh_riworld_2019, riegl_laser_management_systems_gmbh_riprocess_2019, riegl_laser_management_systems_gmbh_rianalyze_2019,riegl_laser_management_systems_gmbh_riprecision_2017}.

The Applanix system allowed post-processing georeferencing of the mobile mapping sensors using the POSPac MMS computational processing software system \cite{trimble_applanix_pospac_2018}.

The photosphere image capture process was based on travel distance rather than time. The field data collection configuration allowed the capture of one image per approximately 10 feet of vehicle traverse travel. In this way, the yielded dataset was spatially balanced and stratified, and allowed for accounting for and remedying temporal variations due to traffic conditions or vehicle stops. The stored original row photospheres image data were converted to JPEG format with an overall width of 8000 pixels, and a height of 4000 pixels. A comma-separated dataset escorted each subset of images (for each field data collection run) that reported for each image capture the \textsl{GNSS} data captured from the Applanix Trimble system.

\subsection{Calibrating and Adjusting for Relative Sensor Positioning}\label{sec3.3}

\begin{figure}[ht]
    \centering
    \includegraphics[width=0.75\textwidth]{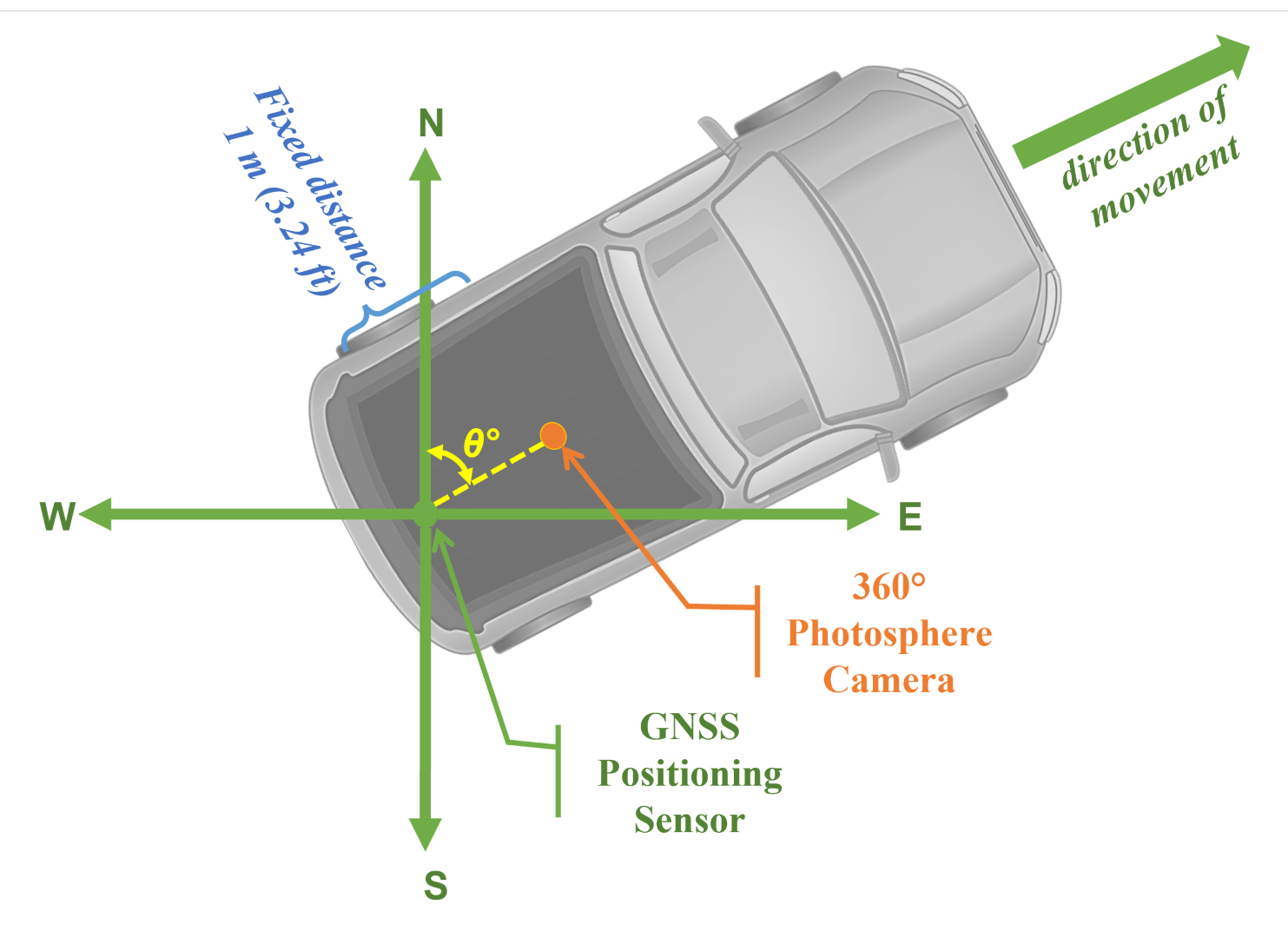}
    \caption{Diagrammatic approximation of the field instrument layout and the angle used for correcting directional GPS and gyroscopic measurements from the positional sensor.}
    \label{fig02}
\end{figure}

Because of the slight deviation in the mobile field data collection unit configuration in terms of the sensor placements, a mathematical/geometric correction in the reporting data was necessary. Specifically, as can be seen in \figurename{~\ref{fig02}}, the relative positioning of True North azimuthal direction of the photosphere camera’s position, and the one obtained from the GPS sensor may differ from each other. This difference is zero when the vehicle travels to the direction of azimuthal True North, in which case the direction of movement is aligned with the central axis of the vehicle’s length (same as the dotted line connecting the GPS and camera sensors in \figurename{~\ref{fig02}}). On the other hand, the deviation increases with the radius of the turn of the vehicle. The more left or right from the True North is the direction of movement, the more different the relative positions are. The distance between the two sensors, $d$ is 1 meter (3.28084 feet). In order to correct for the direction of the movement, we can use the \textit{geometric inverse tangent function} of the angle, expressed in degrees. Symbolically if  $\theta$ is the original difference between the True North axis and the axis of the direction of movement, then the corrected direction $\theta^\prime$ can be calculated from the following logical expressions:
\begin{equation}
    \begin{split}
        \theta = deg(\arctan(\text{Easting},\text{Northing})) - 180\degree \\ 
        \text{if}~ \theta < 0: \theta^\prime = (\theta + 360\degree) \bmod{360} 
    \end{split}
    \label{eq01}
\end{equation}
In other words, the original angle, $\theta$ in degrees can be calculated directly from the Easting and Northing coordinates of the GPS sensor (in State Plane California, Zone 6 Datum Coordinate System \cite{survey_state_1995}) using the inverse tangent function and then converting from radial degrees to directional degrees. If the angle is zero, then $\theta^\prime = \theta$, otherwise, the corrected angle $\theta^\prime$ will be the modulus of the sum of $\theta + 360\degree$ and $360\degree$.

\begin{figure}[ht]
\centering
\includegraphics[width=0.8\textwidth]{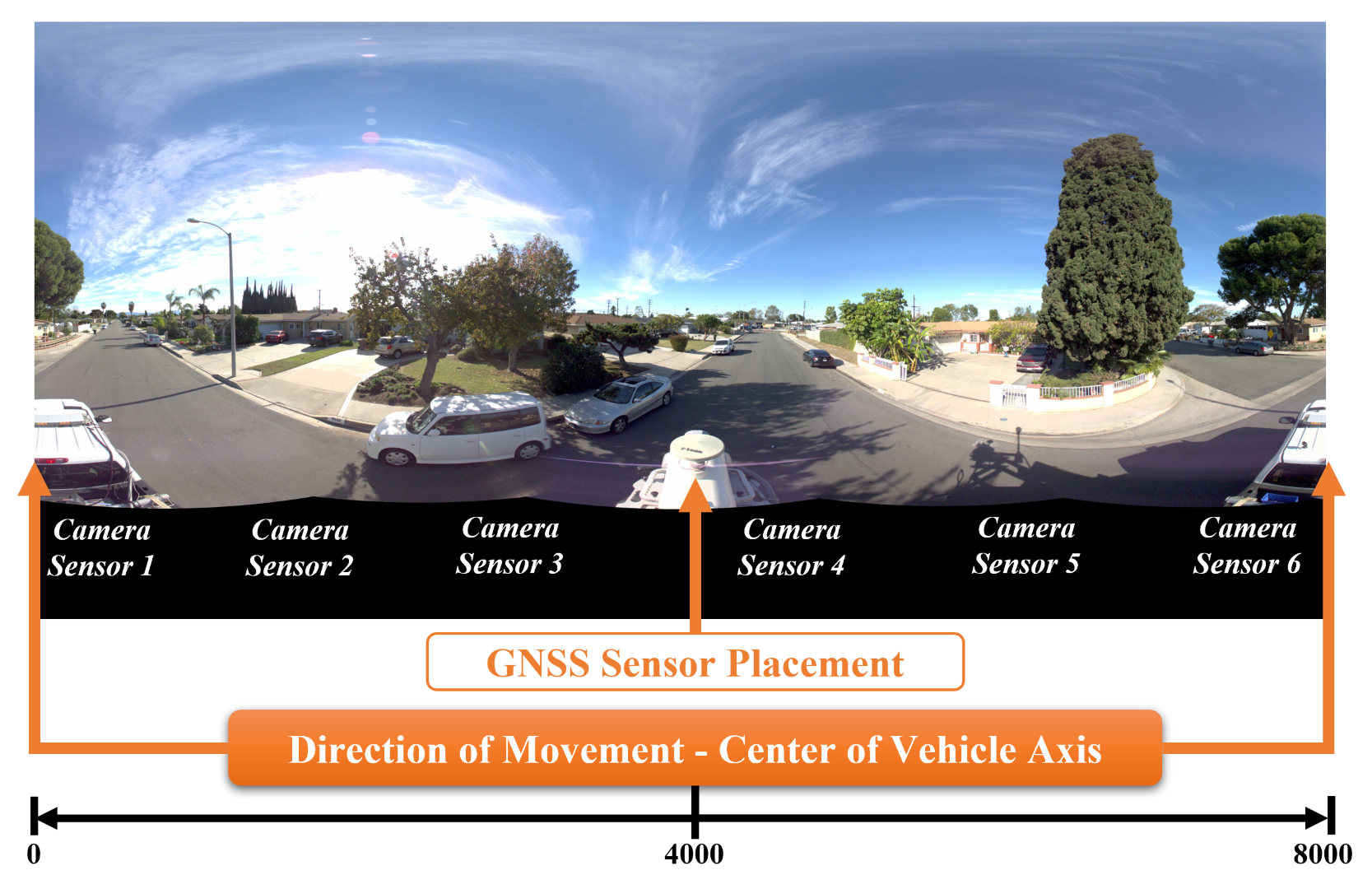}
\caption{The direction of movement corresponds with pixels 1 and 8000 across the width of the sample photosphere image. The sensor can be seen on the middle of the photosphere (i.e., pixel with horizontal width number of 4000). Thus, the direction of movement calculations must be adjusted 180\textdegree~ left or right from the GPS sensor (positive or negative).}
\label{fig03}
\end{figure}

A second calibration that needs to be applied during the calculation process of the photosphere imagery is the relative direction of each horizontal center axis of the image. As mentioned before the flattened photosphere images span over 8000 pixels wide, with the direction of movement corresponding to the first and last horizontal width pixel coordinates (see \figurename{~\ref{fig03}}).  Thus, since the photospheres reflect a 360\textdegree~view of the location, then the derived pixel degree correspondence value is:
\begin{equation}
    \theta_{pixel} = \frac{360\degree}{8000 pt} = 0.045\degree/pt
    \label{eq02}
\end{equation}
Thus, for any pixel $i$ of the image with horizontal location $w_i$, given the direction of movement $\theta_0$, the direction of the line of sight of the pixel can be calculated as:
\begin{equation}
    \theta_i = (\theta_0 + w_i\cdot 0.045) \bmod{360}
    \label{eq03}
\end{equation} 

\subsection{Database and Image Analysis Pre-Processing Methodology} \label{sec3.4}

The original 360\textdegree~equirectangular panorama photosphere imagery is transferred to an Azure blob storage database (cold storage) for processing using \textsl{Azure Python API} libraries \cite{microsoft_azure_azure_2020} and occupy approximately 265 Gb of cold data storage (see also \tablename{~\ref{table01}}). 

\begin{table}[ht]
    \begin{center}
        \begin{minipage}{\textwidth}
            \caption{Blob Containers and Database Sizes by Dataset}\label{table01}
            \begin{threeparttable}
                \begin{tabular*}{\textwidth}{@{\extracolsep{\fill}}ccrrrrrr@{\extracolsep{\fill}}}
                    \toprule
                    Area\tnote{a} & ID & \multicolumn{2}{c}{Photospheres} & \multicolumn{2}{c}{Cardinals} & \multicolumn{2}{c}{Total} \\
                    \cline{3-8}
                    & & Count & Size\tnote{b} & Count & Size\tnote{b} & Count & Size\tnote{b} \\
                    \midrule
                    &337p1 &11,239	&14.85	&89,912		&10.04	&101,151	&24.89\\
                    A&338p1 &3,963	&5.02	&31,704		&3.44	&35,667	&8.46\\
                    &338p2 	&12,145	&15.79	&97,155		&10.73	&109,300	&26.53\\
                    \hline
                    &088p2 	&19,053	&42.06	&152,374	&18.88	&171,427	&60.94\\
                    &089p1 	&12,637	&18.06	&101,076	&12.29	&113,713	&30.35\\
                    &090p1 	&4,936	&7.28	&39,488		&5.00	&44,424	&12.27\\
                    &090p2 	&3,007	&4.15	&24,045	&2.78	&27,052	&6.94\\
                    B&090p3 	&12,261	&18.80	&98,071	&12.80	&110,332	&31.60\\
                    &091p1 	&5,073	&7.43	&40,582	&5.02	&45,655	&12.45\\
                    &091p2 	&6,365	&9.67	&50,908	&6.65	&57,273	&16.62\\
                    &092p1	&10,667	&17.61	&85,320	&11.99	&95,987	&29.60\\
                    &093p1	&1,555	&2.42	&12,440	&1.54	&13,995	&3.96\\
                    \hline
                    &Total	&102,901	&163.45		&823,075	&101.16	&925,976	&264.61\\
                    \bottomrule
                \end{tabular*}
                \begin{tablenotes}
                    \item[a]{\textit{Area:} A: Anaheim Hills; B: North Tustin}
                    \item[b]{\textit{Size:} All sizes measured in gigabytes (Gb), and calculated in Azure blob cold storage.}
                \end{tablenotes}
            \end{threeparttable}
        \end{minipage}
    \end{center}
\end{table}

\begin{figure}[ht]
\centering
\includegraphics[width=0.75\textwidth]{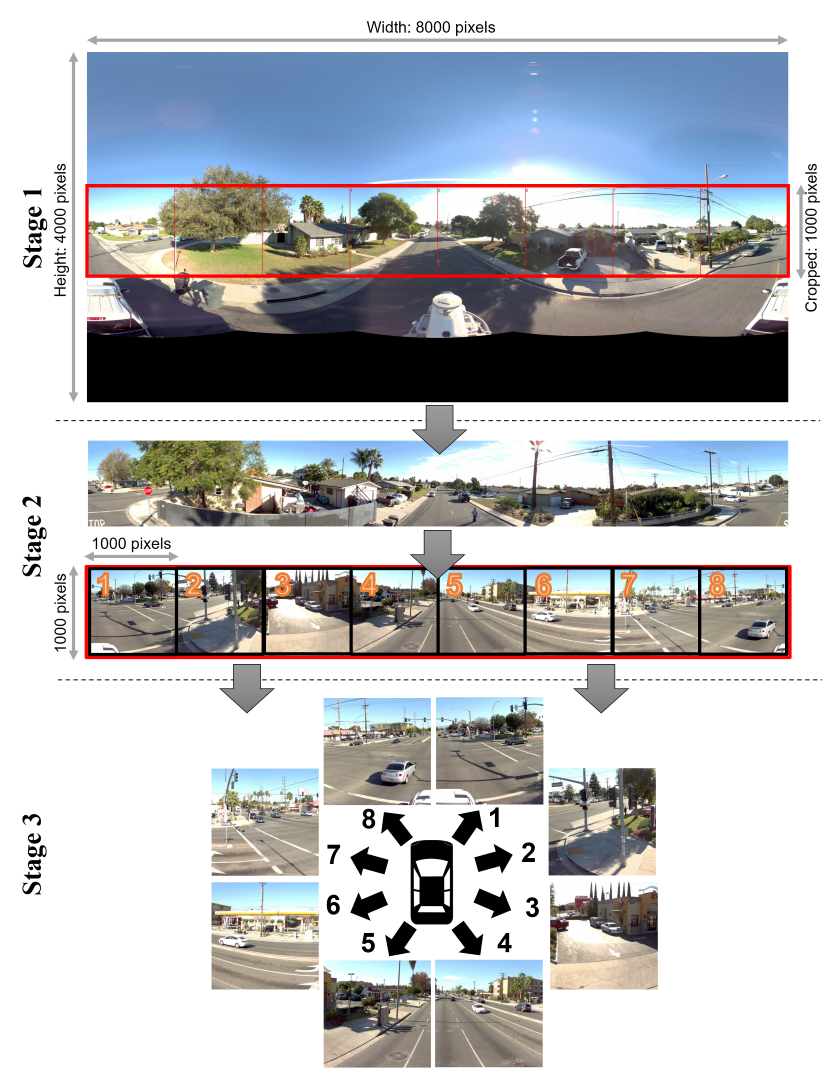}
\caption{A three-stage processing of 360\textdegree~ photosphere images: (a) Stage 1, cropping original photosphere image (4000\texttimes8000 pixels) to a functional area (1000\texttimes8000); (b) Stage 2, separating functional areas into 8 cardinal directions (1000\texttimes1000 pixels each), and; (c) Stage 3, performing analysis of each cardinal images using Azure Cognitive Services Computer Vision API.}
\label{fig04}
\end{figure}

As it will be described further below, images stored as cold blobs allow for simultaneous, and associative storage of custom metadata fields for each of the stored blobs, thus enabling linking sensor data with images. These blob images are processed in three consecutive stages in order to be ready for analysis (see also \figurename{~\ref{fig04}}).

\textbf{Stage 1} involves dynamically and programmatically cropping from the original photosphere image (4000\texttimes8000) a functional analysis area around the image height center (1000\texttimes8000). The reasons for extracting a subset of the original images are multiple. First, it allows for minimizing vertical fish-eye lens distortion of the original camera sensor. Second, minimizes noise and error from detection-irrelevant sections of the image, since the top and bottom part of each photosphere contain sky and blind spot areas respectively. Third, serves the purpose of improving detection accuracy of the deep learning and detection algorithm used for the analysis. In preliminary model runs and experimentation we found that the same model (convoluted ANN) performs better with the reduced-size functional area compared to the original photosphere image. Finally, it systematically and consistently reduces model training and prediction processing times, which is critical in achieving near real-time processing of the ML model.

\textbf{Stages 2 and 3} involve programmatically separating the extracted functional area into eight cardinal sub-images reflecting relevant cardinal directions to the direction of movement. Each of these cardinal images has dimensions of 1000\texttimes1000 pixels. The reason for separating these cardinal images is twofold: a procedural, and an analytical one. 

The procedural reason was derived empirically from subsequent trials and model runs on the photosphere images (and their reduced-from functional area equivalents). Our experiments showed that detection is vastly improved when computer vision object recognition is performed on the smaller than on the wider, and bigger resolution images. The wider photospheres cover a bigger focal area, with multiple potential and candidate objects to be detected, thus both the simultaneous detection probabilities and error rates increase with the size and magnitude of the image to be processed. Furthermore, it appears that both standard and custom models in Microsoft Azure (Azure Cognitive Services Computer Vision/Custom Vision models), and in Google Services (Google Cloud Vision/Custom Vision models) are trained and perform better in images with few, or single focal objects to be detected, rather than multiple object detections in the same image.

The analytical reason for separating these images has to do with the spatial configuration and the nature of objects sought for detection. For example, traffic will always appear in the top right cardinal image (cardinal ID 1), i.e., always immediately to the right of the driving direction (for right-driving roads). Thus, in order to improve model accuracy, minimizing irrelevant and noisy data and improve recognition and detection the cardinal separation serves a very useful purpose. The following \tablename{~\ref{table02}} demonstrates the use of eight nominal cardinal directions if we begin (driving direction) from True North in the NAD83 coordinate system.

\begin{table}[ht]
    \begin{center}
        \begin{minipage}{0.6\textwidth}
            \caption{Nominal Cardinal Photosphere Separation (from True North)}\label{table02}
            \begin{tabular*}{\textwidth}{@{\extracolsep{\fill}}cllc@{\extracolsep{\fill}}}
                \toprule%
                ID &Code &Description &Range\\
                \midrule
                C1	&NNE	&North Northeast	&0\degree - 45\degree\\
                C2	&ENE	&East Northeast		&45\degree - 90\degree\\
                C3	&ESE	&East Southeast		&90\degree - 135\degree\\
                C4	&SSE	&South Southeast	&135\degree - 180\degree\\
                C5	&SSW	&South Southwest	&180\degree - 225\degree\\
                C6	&WSW	&West Southwest		&225\degree - 270\degree\\
                C7	&WNW	&West Northwest		&270\degree - 315\degree\\
                C8	&NNW	&North Northwest	&315\degree - 360\degree\\
                \bottomrule
            \end{tabular*}
        \end{minipage}
    \end{center}
\end{table}

While the coarse classification showcased in \tablename{~\ref{table02}} serves a useful purpose from the nominal True North case, it does not differentiate enough directions to account for all possible starting driving directions present in the data. In order to provide a more refined and suitable cardinal classification, we generated a cardinal lookup dictionary (in the python class script of the ML application), to account for all possible configurations.

Since the driving direction is fixed with 6 decimal point float accuracy, and we need to generate eight cardinal directions starting from the left to right direction across our image, then we know the exact direction of the center of each of these cardinal images starting with $\theta_1$ = $\theta_0$ + 22.5\textdegree~ for the first image (with $\theta_0$ representing the cardinal direction), and for each of the cardinal images adding another 45\textdegree~(modulo 360), i.e., for each \textit{i}=1,2…,8:
\begin{equation}
    \forall i:
    \begin{cases}
        \text{if $i = 1$,} &\theta_i = (\theta_0 + 22.5\degree) \bmod{360}\\
        \text{else,} &\theta_i = (\theta_{i-1} + 45.0\degree) \bmod{360}\\
    \end{cases}
    \label{eq04}
\end{equation}

Using the center direction of each of these eight cardinal images, we can classify each image in one of the following 32 directional classes each one representing a 11.25\textdegree~ range (since there are two directions from the image’s center, i.e., $2\times11.25\degree=22.5\degree$). The cardinal lookup dictionary directions used for the analysis are shown in \tablename{~\ref{table03}}. These lookup values are used also for naming conventions of the cardinal photosphere images, and for visualizing results in spatial (GIS) and non-spatial applications post-processing.

\begin{table}[ht]
    \begin{center}
        \begin{minipage}{0.7\textwidth}
            \caption{Real Cardinal Separation (from Driving Direction)}\label{table03}
            \begin{tabular*}{\textwidth}{@{\extracolsep{\fill}}lllc@{\extracolsep{\fill}}}
                \toprule
                ID &Code &Description &Range\\
                \midrule
                D1	&N		&North				&354.375\degree - 5.625\degree\\
                D2	&NbE	&North by East		&5.625\degree - 16.875\degree\\
                D3	&NNE	&North Northeast	&16.875\degree - 28.125\degree\\
                D4	&NEbN	&Northeast by North	&28.125\degree - 39.375\degree\\
                D5	&NE		&Northeast			&39.375\degree - 50.625\degree\\
                D6	&NEbE	&Northeast by East	&50.625\degree - 61.875\degree\\
                D7	&ENE	&East Northeast		&61.875\degree - 73.125\degree\\
                D8	&EbN	&East by North		&73.125\degree - 84.375\degree\\
                D9	&E		&East				&84.375\degree - 95.625\degree\\
                D10	&EbS	&East by South		&95.625\degree - 106.875\degree\\
                D11	&ESE	&East Southeast		&106.875\degree - 118.125\degree\\
                D12	&SEbE	&Southeast by East	&118.125\degree - 129.375\degree\\
                D13	&SE		&Southeast			&129.375\degree - 140.625\degree\\
                D14	&SEbS	&Southeast by South	&140.625\degree - 151.875\degree\\
                D15	&SSE	&South Southeast	&151.875\degree - 163.125\degree\\
                D16	&SbE	&South by East		&163.125\degree - 174.375\degree\\
                D17	&S		&South				&174.375\degree - 185.625\degree\\
                D18	&SbW	&South by West		&185.625\degree - 196.875\degree\\
                D19	&SSW	&South Southwest	&196.875\degree - 208.125\degree\\
                D20	&SWbS	&Southwest by South	&208.125\degree - 219.375\degree\\
                D21	&SW		&Southwest			&219.375\degree - 230.625\degree\\
                D22	&SWbW	&Southwest by West	&230.625\degree - 241.875\degree\\
                D23	&WSW	&West Southwest		&241.875\degree - 253.125\degree\\
                D24	&WbS	&West by South		&253.125\degree - 264.375\degree\\
                D25	&W		&West				&264.375\degree - 275.625\degree\\
                D26	&WbN	&West by North		&275.625\degree - 286.875\degree\\
                D27	&WNW	&West Northwest		&286.875\degree - 298.125\degree\\
                D28	&NWbW	&Northwest by West	&298.125\degree - 309.375\degree\\
                D29	&NW		&Northwest			&309.375\degree - 320.625\degree\\
                D30	&NWbN	&Northwest by North	&320.625\degree - 331.875\degree\\
                D31	&NNW	&North Northwest	&331.875\degree - 343.125\degree\\
                D32	&NbW	&North by West		&343.125\degree - 354.375\degree\\
                \bottomrule
            \end{tabular*}
        \end{minipage}
    \end{center}
\end{table}

\subsection{Automated Production Workflows} \label{sec3.5}

\begin{figure}[ht]
    \centering
    \includegraphics[width=0.8\textwidth]{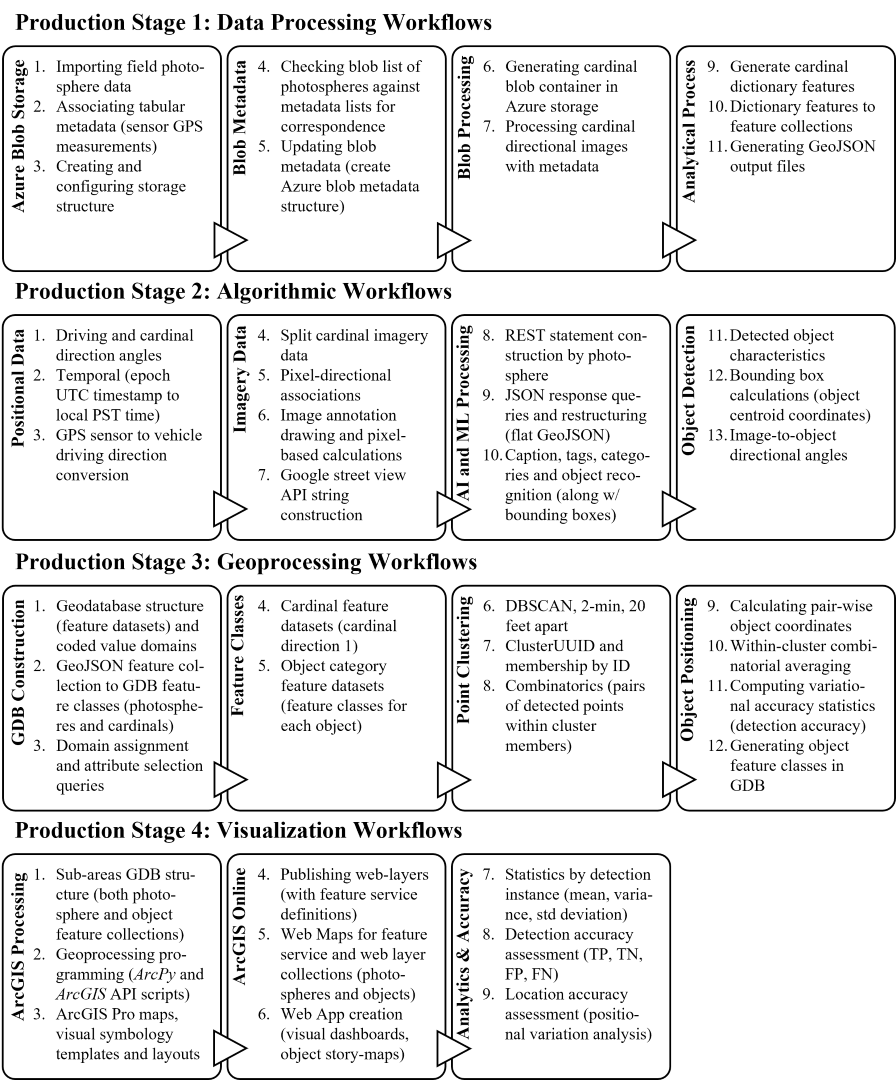}
    \caption{Symbolic diagram for multiple production workflow stages: (1) data processing workflows; (2) algorithmic workflows; (3) geoprocessing workflows, and; (4) visualization workflows.}
    \label{fig05}
\end{figure}

We automated the entire production workflow and processing of the field data acquisition using Python programming. We designed four distinct and associated production workflow stages. These stages are shown diagrammatically in \figurename{~\ref{fig05}} and are summarily described below.

\textbf{Production stage 1}~\textsl{(data processing workflows)} involves configuring and staging Azure blob storage and metadata operations, including custom medatada fields that hold ML object recognition output variables along with field sensor data variables. These operations are followed by analytical processing operations on cardinal image processing (see detailed process on the methodology subsections \ref{sec3.2}, \ref{sec3.4} and \ref{sec3.7} below) and compiling structured \textsl{geoJSON} dictionaries from deep learning custom vision operations.

\textbf{Production stage 2}~\textsl{(algorithmic workflows)} involves calculating and processing positioning spatio-temporal data from obtained sensor information; processing imagery data for cardinal and pixel-directional associations, annotations, and other string construction operations; multiple operations related to AI and ML processing algorithmic tasks (\textsl{REST} construction, \textsl{JSON} response dictionary construction, object tagging and categorization, etc.), and post-AI object detection calculation (triangulating object position, object centroid calculation, and image-to-object distance and angles) as it is described in detail in subsequent sessions.

\textbf{Production stage 3}~\textsl{(geoprocessing workflows)} programmatically follows the AI algorithmic object extraction stage and involves generating and constructing positional data and detected object geodatabases from \textsl{geoJSON} dictionary responses, including feature collections and feature classes with positional and ML detected object-related attributes. In this stage, feature class data are separated into feature datasets by cardinal direction, object category, and object locations, processes that are followed by statistical clustering analysis (\textsl{DBSCAN}) and combinatorial positional calculations for multiple detections. The final operations in this stage involve pair-wise object coordinate extraction, and within-cluster combinatorial averaging and post-processing variational accuracy assessment.

\textbf{Production stage 4}~\textsl{(visualization workflows)} is the final stage that involves programmatically processing spatial geodatabases and feature datasets through \textsl{ArcGIS} \textsl{arcPy} programming classes, and \textsl{ArcGIS API} processing for constructing spatial \textsl{REST} feature services, web maps, and web apps used in public data portal, along with analytics and spatial accuracy detection accuracy metrics. All four production stages are fully automated and are processed by dataset, in such way that the processing that begins from data acquisition caries through final data portal production and visualization.

\begin{figure}[ht]
    \centering
    \includegraphics[width=0.75\textwidth]{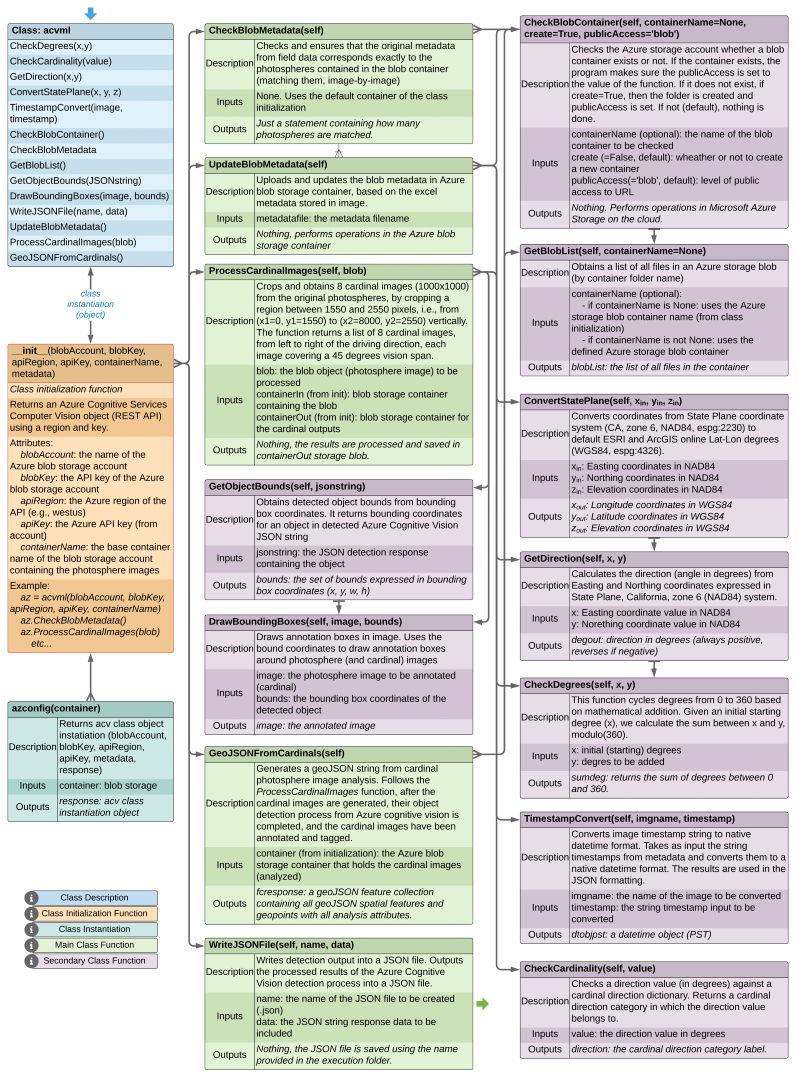}
    \caption{Symbolic class diagram for the main data processing Python function \textsl{\texttt{acvml}}, implementing photosphere image processing, object recognition, attribute extraction and tagging.}
    \label{fig06}
\end{figure}

\begin{figure}[ht]
    \centering
    \includegraphics[width=0.75\textwidth]{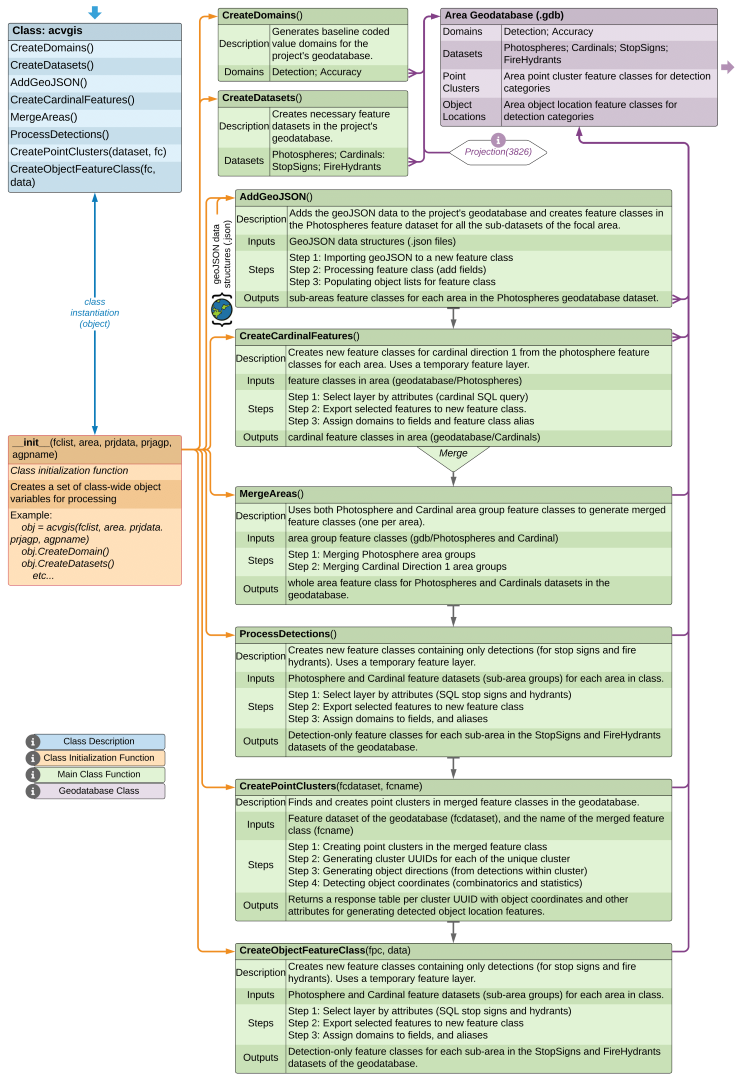}
    \caption{Symbolic class diagram for the spatial geoprocessing Python function \textsl{\texttt{acvgis}}, implementing geodatabase and feature class construction, generating feature services and \textsl{ArcGIS} online datasets.}
    \label{fig07}
\end{figure}

Two distinct python classes were designed and implemented for the production workflows described here. Specifically, the main data processing class \textsl{\texttt{acvml}}, shown using the class diagram of \figurename{~\ref{fig06}}, and the spatial geoprocessing class acvgis, shown in the class diagram of \figurename{~\ref{fig07}}. 

Both Python classes contain multiple functions for sequential data processing, and the output \textsl{geoJSON} file of \textsl{\texttt{acvml}} is used in processing spatial feature classes and geodatabase operations in \textsl{\texttt{acvgis}}.

Class initialization and object instantiation is performed by sub-dataset (blob storage container) of the field data, and the automated production process adds the deep learning object recognition results at multiple outputs: (a) as metadata at each cardinal blob of photosphere images in the blob storage containers; (b) as \textsl{geoJSON} dictionary class members; and (c) as attributes in geodatabase feature classes and resulting features in \textsl{ArcGIS} online \textsl{REST} feature services.

\subsection{Machine Learning and Computer Vision Methods}\label{sec3.6}

We used and tested multiple ML algorithms for Computer Vision. Specifically, we tested in production data:
\begin{enumerate}[a.]
    \item{Microsoft Azure}
    \begin{enumerate}[i.]
        \item{\textsl{Cognitive Services Vision API}}
        \item{\textsl{Cognitive Services Custom Vision API}}
    \end{enumerate}
    \item{Google Cloud Services}
    \begin{enumerate}[i.]
        \item{\textsl{Cloud Vision API}}
        \item{\textsl{Custom Vision API}}
    \end{enumerate}
    \item{Mathematica \textsl{Image Processing} Vision Model}
    \item{Custom Python model using \textsl{\texttt{TensorFlow}} and \textsl{\texttt{Keras}} algorithms}
\end{enumerate}

\begin{figure}[ht]
    \centering
    \includegraphics[width=0.75\textwidth]{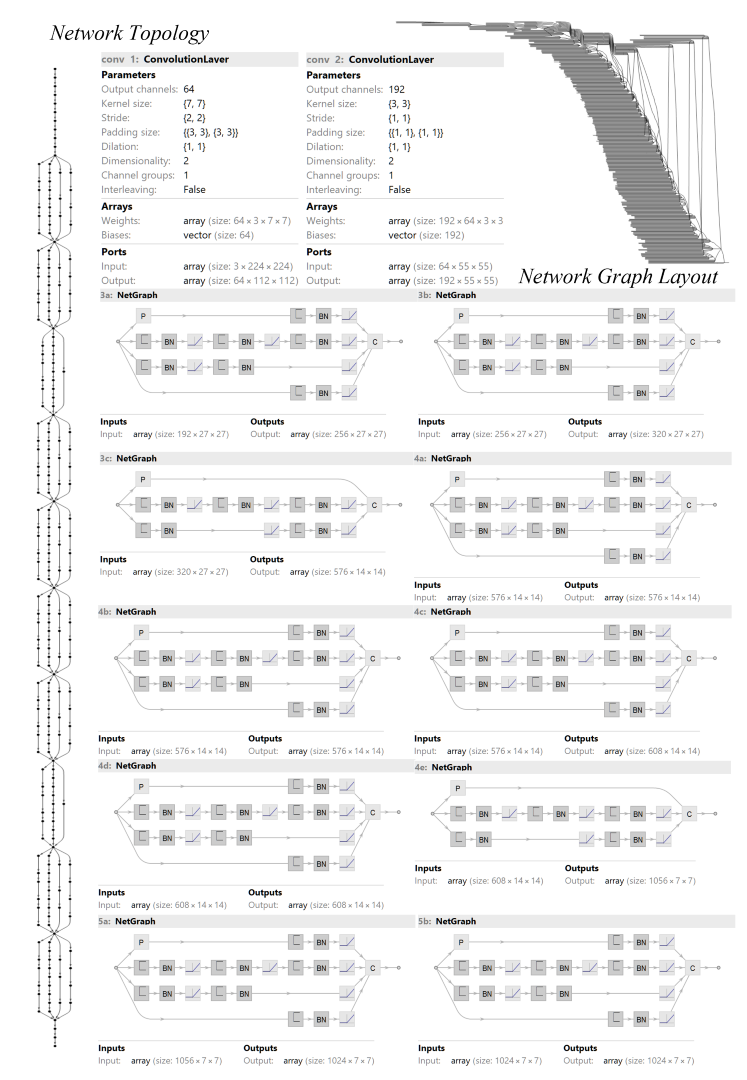}
    \caption{The topology and configuration of the trained deep learning model reflecting a convoluted artificial neural network, as trained in Mathematica.}
    \label{fig08}
\end{figure}

From these models, (a.i), (b.i) and (c) used object classifiers pre-trained through the convolution deep neural network models from web images. An example of the topological configuration of the Mathematica's Image Processing model showcasing the layout of the convolution layers and network in the deep neural network model is shown in \figurename{~\ref{fig08}}. We used classification results from these models to perform additional, custom classifications to train models (a.ii) and (b.ii). Specifically, we assessed the accuracy of object classification obtained by the Azure's Cognitive Services Vision model and Google Cloud Vision model, by identifying images correctly classified by the models, and using approximately 50-100 of these images (depending of the object: 100 for stop signs, 50 for fire hydrants) for training the custom vision version of the models in (a) through (d).

\subsection{Detected Object Position Calculation}\label{sec3.7}

The relationship between a slope of a line ($\mu$) and its angle in degrees ($\theta\degree$) is:
\begin{equation}
    \tan\theta = \mu
    \label{eq05}
\end{equation}
and the slope between two coordinate points is,
\begin{equation}
    \mu = \frac{y_2 - y_1}{x_2 - x_1}
    \label{eq06}
\end{equation}

We have two GPS points (from a vehicle) that are known, i.e., point $A = (x_A, y_A)$, and $B = (x_B, y_B)$. We want to find the unknown coordinates of a detected object, $C = (x_C, y_C)$. For such object, we only know its detected directions (angles) from points $A$ and $B$ respectively, namely $\hat{\theta}_A$ and $\hat{\theta}_B$.

We know that the two lines intersect at point $C$. Therefore, from the slope equations, we have,
\begin{equation}
    \begin{split}
        \mu_A = \tan\theta_A = \frac{y_A - y_C}{x_A - x_C}\\
        \mu_B = \tan\theta_B = \frac{y_B - y_C}{x_B - x_C}
    \end{split}
    \label{eq07}
\end{equation}
then,
\begin{equation}
    \begin{split}
        y_C = y_A - \tan\theta_A\cdot(x_A - x_C)\\
        y_C = y_B - \tan\theta_B\cdot(x_B - x_C)
    \end{split}
    \label{eq08}
\end{equation}
and, dividing the two equations in (\ref{eq08}), we have,
\begin{equation}
    y_A-\tan\theta_A\cdot(x_A-x_C) = y_B-\tan\theta_B\cdot(x_B-y_B)
    \label{eq09}
\end{equation}
and therefore,
\begin{equation}
    x_C=\frac{y_B-y_A+\tan\theta_A\cdot x_A-\tan\theta_B\cdot x_B}{\tan\theta_A-\tan\theta_B}
    \label{eq10}
\end{equation}
or, finally,
\begin{equation}
    \begin{split}
        x_C=\frac{y_B-y_A+\mu_A\cdot x_A-\mu_B\cdot x_B}{\mu_A-\mu_B}\\
        y_C=y_A-\mu_A\cdot(x_A-x_C)
    \end{split}
    \label{eq11}
\end{equation}

The pair of values $x$ and $y$ in equation \ref{eq11} represent the two GPS coordinates for point $C$ expressed in \textsl{State Plane California (zone 6)} coordinate system \cite{survey_state_1995} (Easting and Northing).

Once all three positions are known, $A$, $B$, and $C$, we can calculate also all relevant distances, since,
\begin{equation}
    \begin{split}
        d_{AC}=\sqrt{(x_C-x_A)^2+(y_C-y_A)^2}\\
        d_{BC}=\sqrt{(x_C-x_B)^2+(y_C-y_B)^2}
    \end{split}
    \label{eq12}
\end{equation}

\begin{figure}[ht]
    \centering
    \includegraphics[width=\textwidth]{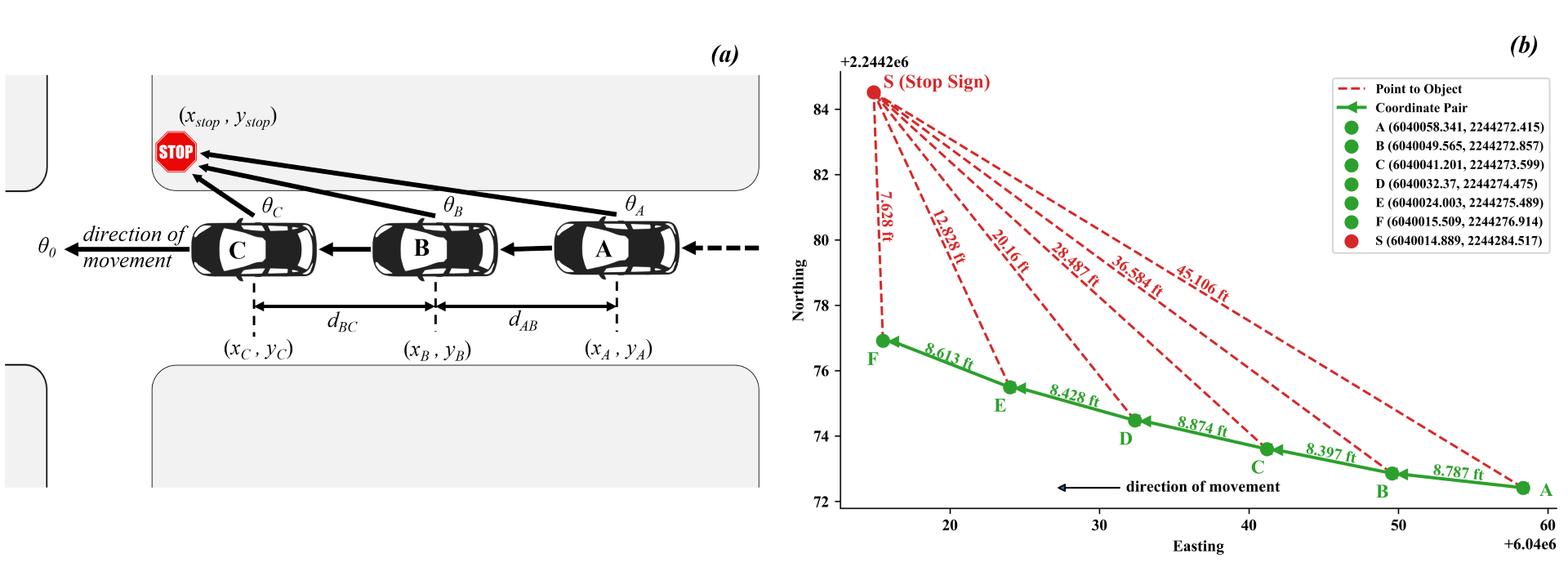}
    \caption{Example visualization of positional and distance calculation of object location: subgraph (a): sequential detection properties; subgraph (b) real-data example of stop sign detection, from the six consecutive positional data captures, A-F, along the direction of movement (in NAD83 coordinate system).}
    \label{fig09}
\end{figure}

\subsection{Spatial Clustering of Object Detection}\label{sec3.8}

Following the object identification process using the data processing (stage 1), algorithmic (stage 2) and early stages of geoprocessing workflows (GDB construction and feature dataset compilation of stage 3) described in section \ref{sec3.5} and \figurename{~\ref{fig05}}, a spatial clustering process is required to uniquely identify the location of each object. In summary, the process (a) heuristically identifies detection groups with statistical likelihood of referencing a single unique object and clusters them into a cluster group, and; (b) uses pair-wise combinatorial statistics for distance calculations (see also section \ref{sec3.7} above) by calculating the statistical mean and standard deviation of object positioning coordinates.

The use of spatial clustering algorithms has been well documented in the relevant geospatial literature and has been among the most powerful tools in the geostatistical toolset of analysis.

\section{Results}\label{sec4}

The analytical results obtained from the experimental and methodological processing described in the previous section are provided in the following subsections. For this experimental application, we used two unincorporated areas in Orange County, covering a total of 45 square miles, and traversing a total of 195 miles collecting sample observation data (photospheres and 3D \textsl{LiDAR} imagery) every 10 feet for both driving directions. We consequently processed the data according to the methodology described in the previous section and obtained object detection and locational estimations according to the implemented algorithms. We present the analysis of two types of objects: stop signs (1,562 detected) and fire hydrants (425 detected) with specific examples integrating 3D \textsl{LiDAR} imagery data. Finally, we provide an accuracy assessment of the detection and location estimates for the sampled object data.

\subsection{Geographic Coverage in Collection Areas}\label{sec4.1}

\begin{figure}[ht]
    \centering
    \includegraphics[width=\textwidth]{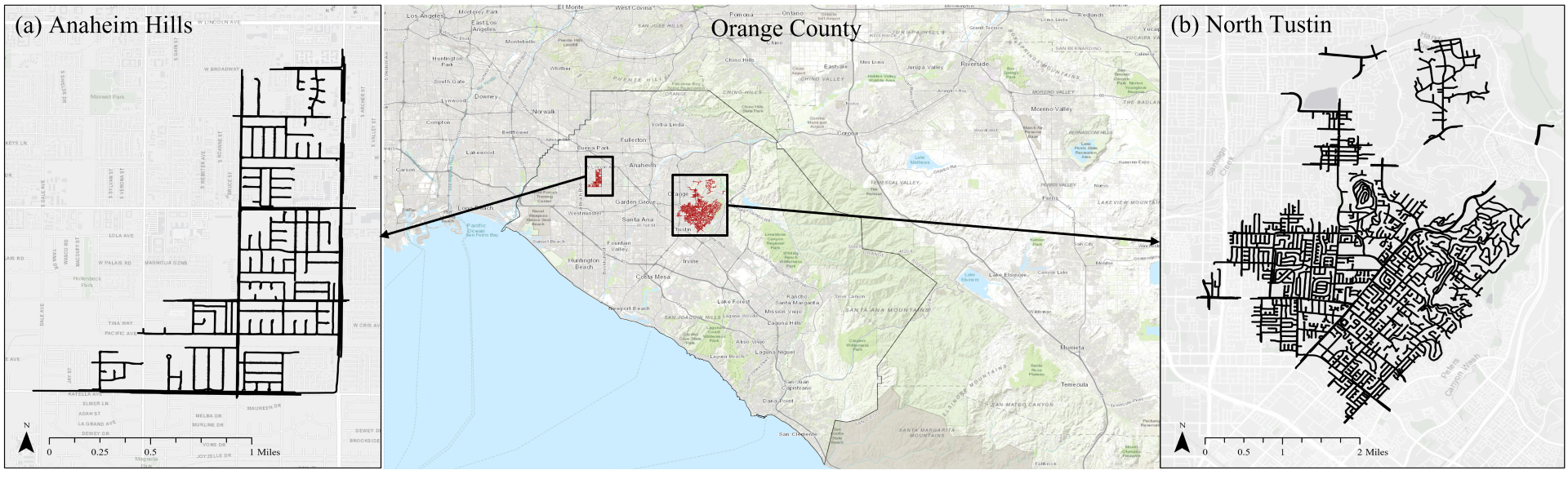}
    \caption{An overview of the data collection areas in Orange County, California: (a) Anaheim Hills, approximately 5 square miles of coverage area, and (b) North Tustin, approximately 40 square miles of coverage area.}
    \label{fig10}
\end{figure}

As can be seen in \figurename{~\ref{fig10}}, the two collection areas are in the north-central part of Orange County. These areas are the Anaheim Hills, and the North Tustin areas, both part of the County of Orange unincorporated areas.

The two sample collection areas used for the classification cover approximately a combined 45 square mile area as it can be measured using an approximation of a minimum confined rectangle containing all the field observations in each of the two regions. As can be seen in \tablename{~\ref{table04}}, the Anaheim Hill unincorporated area covers approximately 5 square miles, while the North Tustin area covers approximately 40 square miles. In terms of the amount of collected photospheres, a total of 102,901 360\textdegree~images (8,000×1,000 pixels each) were collected and processed, of which 27,347 photospheres were collected in the Anaheim Hills area, and 75,554 photospheres were collected in the North Tustin area. From these 360\textdegree~images, using the processing algorithmic methodologies described in the methodology section, we obtained and processed successfully a total of 822,492 cardinal photosphere images (1,000×1,000 pixels each) for both collection areas: 218,389 cardinal images for the Anaheim Hills area, and 604,103 cardinal images for the North Tustin area.

Given the approximate 10 feet distance between two consecutive data captures of our data collection instruments, we can make an estimation of the distance covered within our collection areas, a total of 194.9 miles travelled, with 51.8 miles covered in the Anaheim Hills area and 143.1 miles covered in the North Tustin area. The summary of the collection statistics for the sample areas can be seen in \tablename{~\ref{table04}}.

\begin{table}[ht] 
    \begin{center}
        \begin{minipage}{0.75\textwidth}
            \caption{Image Classification Results and Counts}\label{table04}
            \begin{threeparttable}
                \begin{tabular*}{\textwidth}{@{\extracolsep{\fill}}lrrr@{\extracolsep{\fill}}}
                    \toprule%
                    Category &Anaheim Hills &North Tustin &Total\\
                    \midrule
                    Photospheres	&27,347		&75,554		&102,901\\
                    Cardinals		&218,389	&604,103	&822,492\\
                    Distance (mi)	&51.8		&143.1		&194.9\\
                    \hline
                    Area (sq. mi)\tnote{a}	&5	&40	&45\\
                    \bottomrule
                \end{tabular*}
                \begin{tablenotes}
                    \item[a]{The area is calculated approximately using a minimum confined rectangle containing all the captured locations in each perspective collection area.}
                \end{tablenotes}
            \end{threeparttable}
        \end{minipage}
    \end{center}
    \end{table}

For the approximately 102,901 cardinal images in the first cardinal direction with respect to the driving direction (top-left) run through our detection algorithms, we found approximately 1,562 stop sign detections, and 425 fire hydrant detections for both data collection sample areas. Specifically, in the Anaheim Hills area (27,347 cardinal directions analyzed), we detected 287 stop signs, and 123 fire hydrants. These correspond to approximately 5.54 stop signs and 2.38 fire hydrants per mile sampled respectively. In the North Tustin area (75,554 cardinal directions analyzed), we detected 1,275 stop signs and 302 fire hydrants, corresponding to 8.91 stop signs and 2.11 fire hydrants per mile sampled respectively. More in-depth algorithmic detection statistics are shown in \tablename{~\ref{table05}}.

\begin{table}[ht]
    \begin{center}
        \begin{minipage}{0.75\textwidth}
            \caption{Spatial Object Detection Frequency and Distance Statistics}\label{table05}
            \begin{threeparttable}
                \begin{tabular*}{\textwidth}{@{\extracolsep{\fill}}llrrrr@{\extracolsep{\fill}}}
                    \toprule%
                    Area &ID &SS\tnote{a} &FH\tnote{b} &SS/mi\tnote{c} &FH/mi\tnote{d}\\
                    \midrule
                    Anaheim&337p1 &81 &94 &3.81 &4.42\\
                    Hills&338p1 &73 &10 &9.73 &1.33\\
                    &338p2 &133 &19 &5.78 &0.83\\
                    \hline
                    &088p2 &364 &142 &10.09 &3.94\\
                    &089p1 &167 &44 &6.98 &1.84\\
                    &090p1 &100 &9 &10.70 &0.96\\
                    North &090p2 &47 &5 &8.26 &0.88\\
                    Tustin &090p3 &174 &29 &7.49 &1.25\\
                    &091p1 &78 &15 &8.12 &1.56\\
                    &091p2 &129 &19 &10.70 &1.58\\
                    &092p1 &198 &37 &9.80 &1.83\\
                    &093p1 &18 &2 &6.12 &0.68\\
                    \hline
                    Total &&1,562 &425 &8.13 &1.76\\	
                    \bottomrule
                \end{tabular*}
                \begin{tablenotes}[para]
                    \item[a]{SS: stop signs}
                    \item[b]{FH: fire hydrants}
                    \item[c]{SS/mi: mean number of stop signs per mile samples}
                    \item[d]{FH/mi: mean number of fire hydrants per mile sampled}
                \end{tablenotes}
            \end{threeparttable}
        \end{minipage}
    \end{center}
\end{table}

\subsection{Example of Stop Sign Detection}\label{sec4.2}

\begin{figure}[ht]
    \centering
    \includegraphics[width=\textwidth]{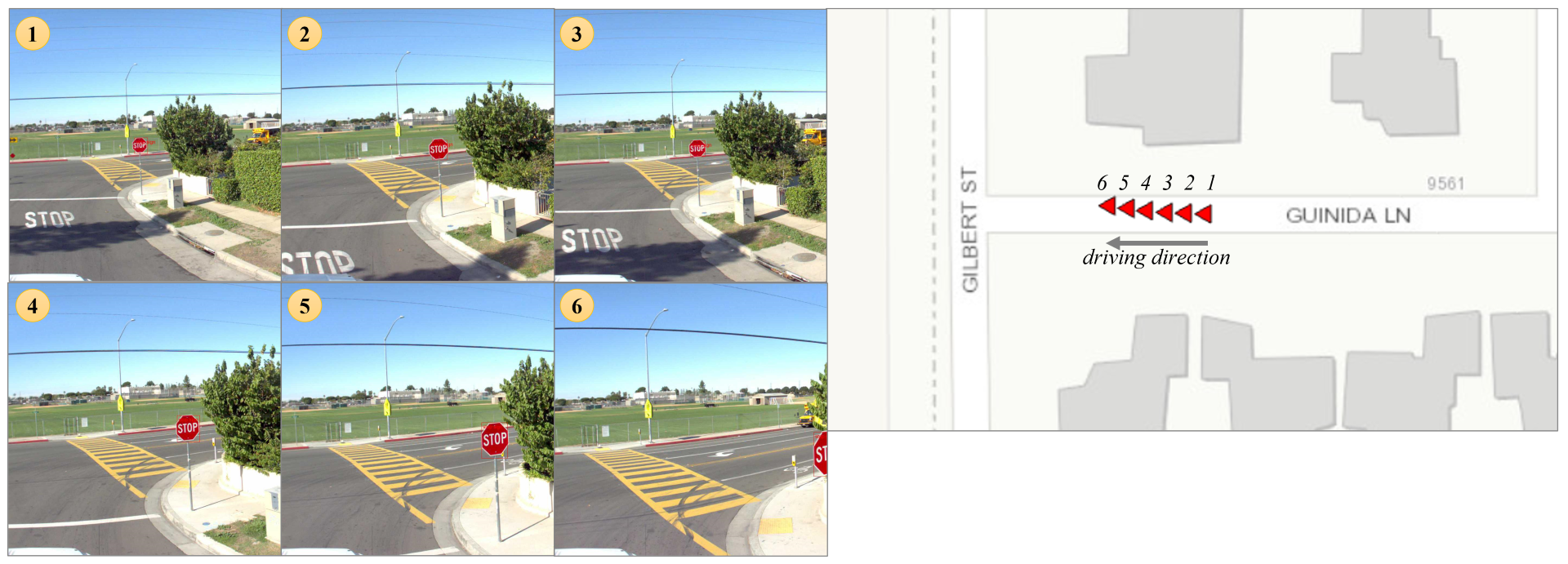}
    \caption{A graphical representation/example of multiple object detections for a single stop sign in Anaheim Hills, Orange County, CA. The six subfigures of the top two image rows, correspond to the detection IDs in the bottom subfigure, with numbering corresponding to sequential detections.}
    \label{fig11}
\end{figure}

An example of multiple detections for a single stop sign can be seen in \figurename{~\ref{fig11}}. Given that photosphere field data captures occurred at least every 10 feet apart we can see that the stop sign was detected at least 70 feet away (the last detection estimated the point-to-object distance, around 14 feet from the sensor).

The estimation statistics and basic accuracy data are provided in \tablename{~\ref{table06}}. While both the driving direction $\theta_0$, and the object direction $\theta_s$, incrementally increase, their absolute difference ($\theta_s - \theta_0$) also increases from 21.69\textdegree~in the first detection, to 43.92\textdegree~in the last detection, indicating a widening angle between the sensor location and the object location (stop sign), as one would expect, and illustrated graphically in \figurename{~\ref{fig09}}. 

\begin{table}[ht]
    \begin{center}
        \begin{minipage}{0.9\textwidth}
            \caption{Example Stop Sign Statistics (ID:181204, Anaheim Hills)}\label{table06}
            \begin{threeparttable}
                \begin{tabular*}{\textwidth}{@{\extracolsep{\fill}}crrrrrr@{\extracolsep{\fill}}}
                    \toprule%
                    N &Lat (N) &Lon (W) &Alt &$\theta_0$ &$\theta_s$ &Conf.\\
                    \midrule
                    1 &33.814440 &117.967288 &31.763 &268.780\degree &290.470\degree &0.750\\
                    2 &33.814441 &117.967317 &31.761 &269.558\degree &293.341\degree &0.725\\
                    3 &33.814442 &117.967345 &31.736 &270.970\degree &297.070\degree &0.736\\
                    4 &33.814444 &117.967374 &31.719 &272.760\degree &302.527\degree &0.778\\
                    5 &33.814447 &117.967401 &31.685 &275.070\degree &310.237\degree &0.854\\
                    6 &33.814450 &117.967429 &31.669 &278.175\degree &322.095\degree &0.736\\
                    \hline
                    $\mu$ &33.8144 &117.9674 & & &305.788\degree\\	
                    $\sigma^2$ &0.000014 &0.000012 & & &5.63004\degree\\		
                    \bottomrule
                \end{tabular*}
                \begin{tablenotes}
                    \item{\textit{Lat/Lon}: Latitude/Longitude (WGS84); \textit{Alt}: altitude (sea level elevation) in feet; $\theta_0$: driving direction; $\theta_s$: object direction (stop sign); \textit{conf}: confidence (model-estimated probability); $\mu$: mean value; $\sigma^2$: standard deviation value.}
                \end{tablenotes}
            \end{threeparttable}
        \end{minipage}
    \end{center}
\end{table}

The mean driving distance of the multi-detection estimation was $\mu_0= 17.15719$ feet, with mean standard deviation of the estimation $\sigma^2_0=10.93862$ feet. The mean estimated distance between points at detection and the detected stop sign object was $\mu_1=13.95209$ feet, with mean standard deviation of $\sigma^2_1=9.146422$ feet.

\begin{figure}[ht]
    \centering
    \includegraphics[width=\textwidth]{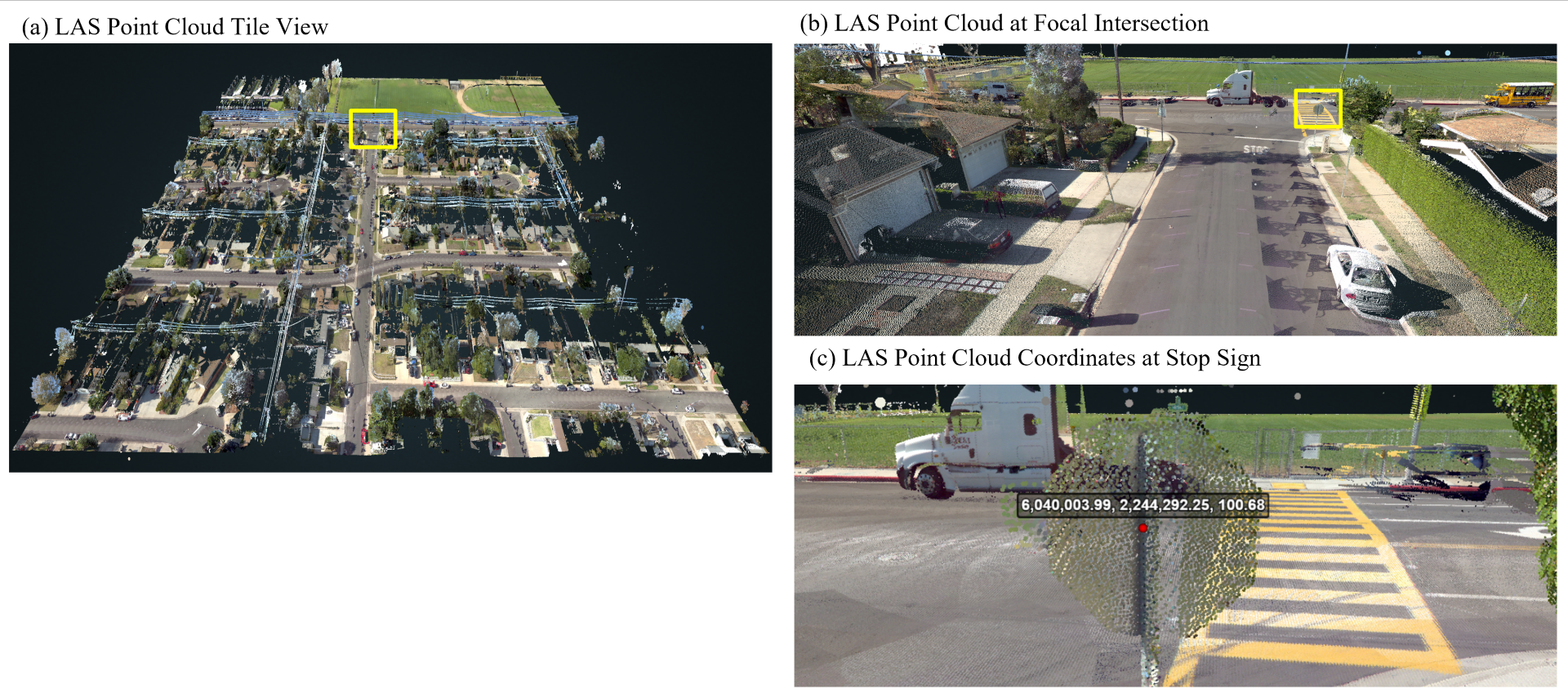}
    \caption{Visualization of LAS point cloud data for the focal example area. The top sub-graph (a) shows the whole tile sub-dataset for the merged area data collection runs. Subgraph (b) provides a zoomed-in view of the focal intersection with the stop sign visible at the end of the road. Subgraph (c) presents the GPS coordinates (NAD83, CA zone 6) of the center of the sign.}
    \label{fig12}
\end{figure}

We have verified the prediction accuracy of our model results using the associated \textsl{LiDAR} point cloud data captured simultaneously with the photosphere imagery. As can be seen in \figurename{~\ref{fig12}}(a), the detected stop sign and area corresponds to an associated LAS point cloud data tile (top subgraph), and more specifically to the street intersection shown in subgraph \figurename{~\ref{fig12}}(b). The visual presents a point budget of 10 million points, visualized through a custom Potree application \cite{wimmer_potree_2016}, using a \textsl{LAS} point cloud feature service. The coordinates obtained from the point cloud data, shown in subgraph \figurename{~\ref{fig12}}(c), are identical with the coordinates estimated by the detection algorithm, thus providing an additional validation of the accuracy of our model estimates.

\subsection{Example of Fire Hydrant Detection}\label{sec4.3}

\begin{figure}[ht]
	\centering
	\includegraphics[width=0.6\textwidth]{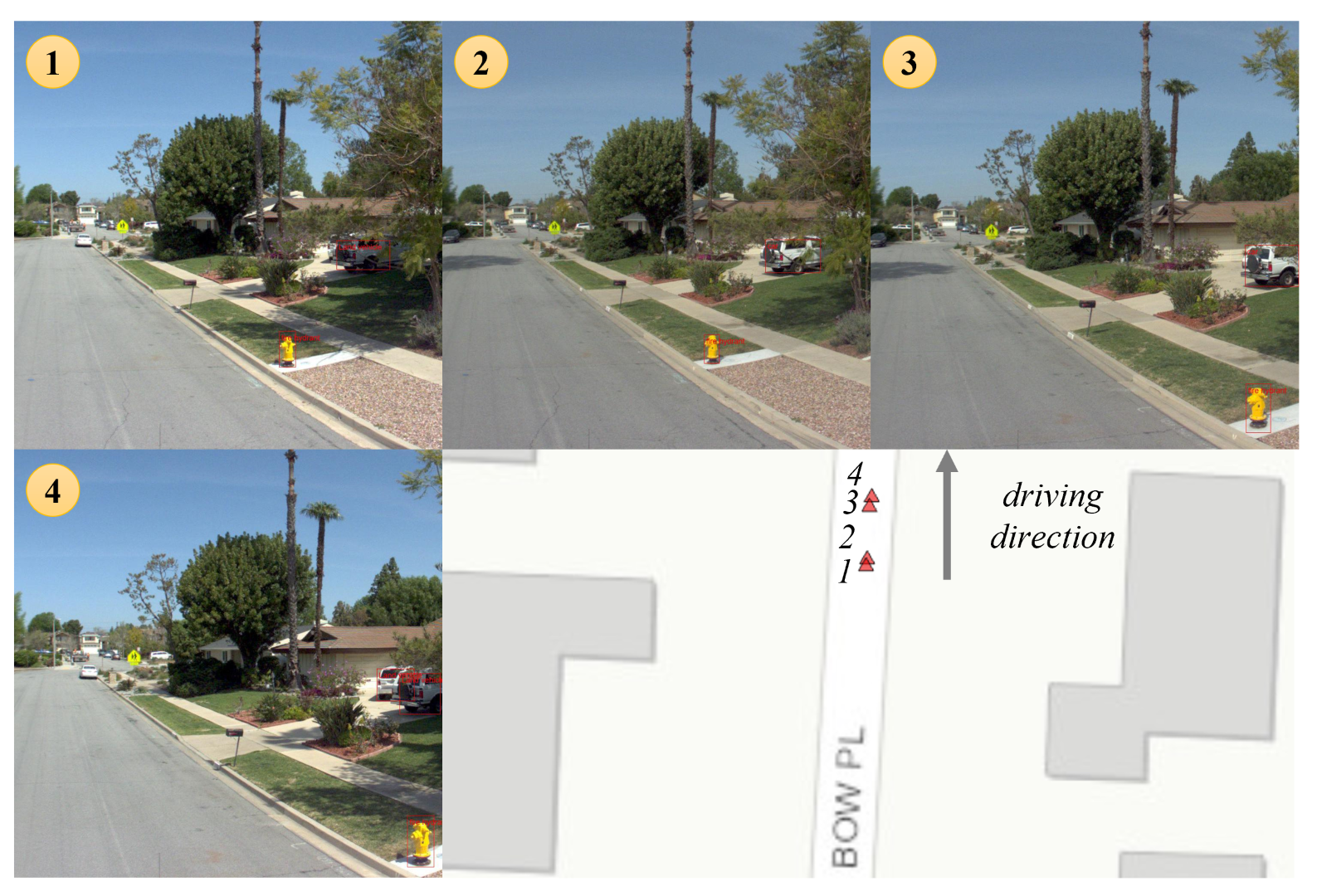}
	\caption{A graphical representation/example of multiple object detections for a single fire hydrant in North Tustin, Orange County, CA. The four subimages, correspond to the detection IDs in the bottom right subfigure, with numbering corresponding to sequential detections.}
	\label{fig13}
\end{figure}

A similar example of multiple detections for a single fire hydrant detection can be seen in \figurename{\ref{fig13}} and the obtained statistics in \tablename{~\ref{table07}}. For this example, we registered and process four detections of a fire hydrant in the North Tustin focal area. The first two detections occur around 50-70 feet away from the fire hydrant, and the last two within 10-30 feet away. As with the stop sign detection example, the  angle between the direction of movement and the detected fire hydrant increases from 5.76\textdegree~ (first detection) to 20.21\textdegree~ (last detection). The estimated mean driving distance among these detections was $\mu_0 = 11.93858$ feet, with mean standard deviation of $\sigma^2_0 = 6.500228$ feet. The estimation of the mean distance between detection points and fire hydrant objects was $\mu_1 = 8.541106$ feet, with mean standard deviation of $\sigma^2_1 = 11.08175$ feet.

\begin{table}[ht]
    \begin{center}
        \begin{minipage}{0.8\textwidth}
            \caption{Fire Hydrant Detection Statistics (ID:190329, North Tustin)}\label{table07}
            \begin{threeparttable}
                \begin{tabular*}{\textwidth}{@{\extracolsep{\fill}}crrrrrr@{\extracolsep{\fill}}}
                    \toprule%
                    N &Lat (N) &Lon (W) &Alt &$\theta_0$ &$\theta_s$ &Conf.\\
                    \midrule
                    1 &33.771314 &117.813695 &69.409 &22.746\degree &28.528\degree &0.532\\
                    2 &33.771319 &117.813694 &69.387 &23.141\degree &29.351\degree &0.636\\
                    3 &33.771357 &117.813692 &69.452 &22.775\degree &40.977\degree &0.704\\
                    4 &33.771364 &117.813690 &69.433 &22.866\degree &43.071\degree &0.609\\
                    \hline
                    $\mu$ &33.77132 &117.8137 & & &36.673\degree\\	
                    $\sigma^2$ &0.000005 &0.000034 & & &3.5723\degree\\	
                    \bottomrule
                \end{tabular*}
                \begin{tablenotes}
                    \item{\textit{Lat/Lon}: Latitude/Longitude (WGS84); \textit{Alt}: altitude (sea level elevation) in feet; $\theta_0$: driving direction; $\theta_s$: object direction (stop sign); \textit{conf}: confidence (model-estimated probability); $\mu$: mean value; $\sigma^2$: standard deviation value.}
                \end{tablenotes}
            \end{threeparttable}
        \end{minipage}
    \end{center}
\end{table}

\begin{figure}[ht]
	\centering
	\includegraphics[width=\textwidth]{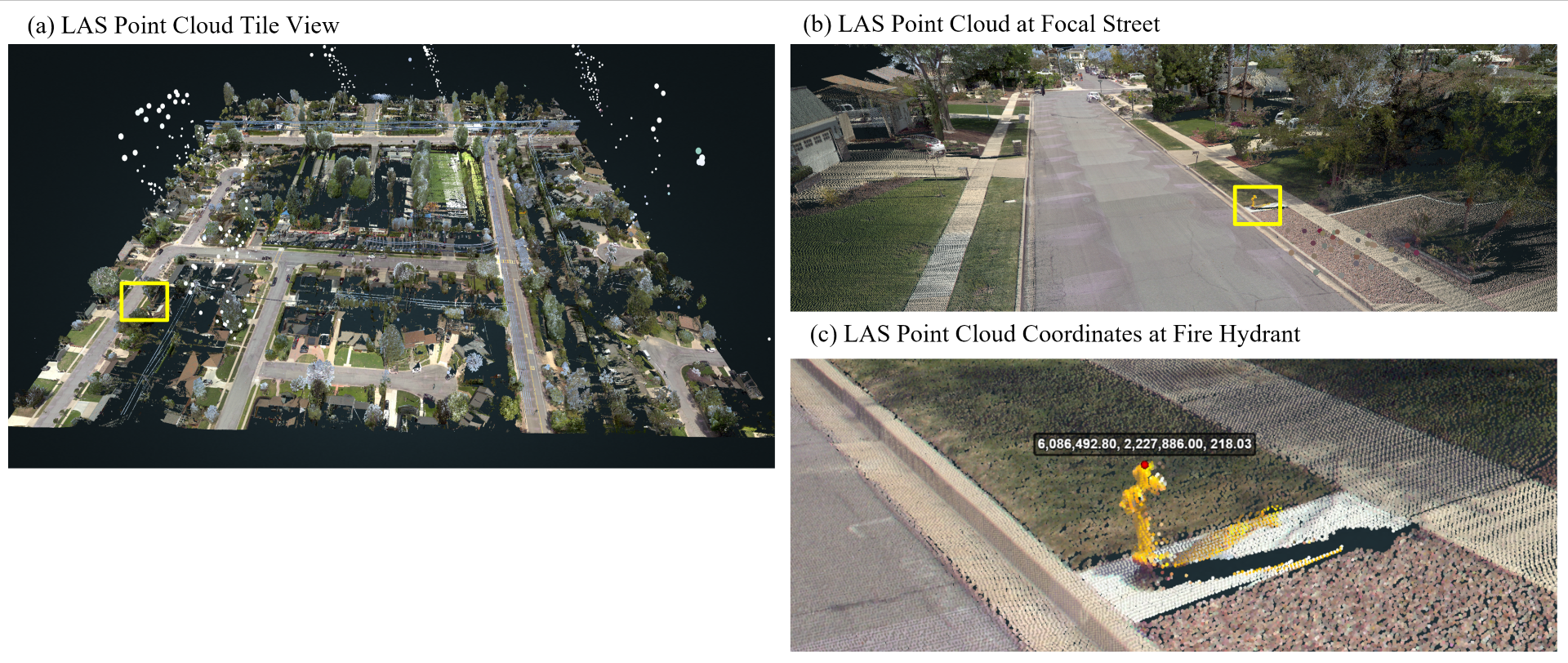}
	\caption{Visualization of LAS point cloud data for the focal example area. The top sub-graph (a) shows the whole tile sub-dataset for the merged area data collection runs. Subgraph (b), provides a zoomed-in view of the focal street with the fire hydrant visible at the right side of the road. Subgraph (c) presents the GPS coordinates (NAD83, CA zone 6) of the fire hydrant.}
	\label{fig14}
\end{figure}

Again, we verified the accuracy of the estimation using the  \textsl{LiDAR} LAS point cloud data for the detected location, and as it can be seen in \figurename{~\ref{fig14}}, the LAS point coordinates in subgraph (c) match the detected and estimated coordinates of our algorithmic approach.

\subsection{Accuracy Assessment of Spatial Object Detection Algorithm}\label{sec4.4}

Beyond the spatial visual and graphical verification of the accuracy of our detection approach in terms of the estimated location coordinates, we also evaluated the estimation accuracy of the estimation. We assessed the variability of the estimates produced by the detection algorithm, and evaluated the mean locational estimate values, along with the variance of their detection. Given the probabilistic nature of the machine learning and computer vision framework we employed for this analysis, one would expect to see variability increased with the detection distance between a driving point and the detected object. This variability is further compounded by the constraints imposed by the physical measurement instrumentation, specifically by the resolution of the imagery used for the estimation.

\begin{table}[ht]
    \begin{center}
        \begin{minipage}{0.8\textwidth}
            \caption{Accuracy Assessment Statistics of Object Location}\label{table08}
            \begin{threeparttable}
                \begin{tabular*}{\textwidth}{@{\extracolsep{\fill}}cccrrrrrr@{\extracolsep{\fill}}}
                    \toprule%
                    Area &Object &N &\multicolumn{3}{c}{$\sigma^2_{lat}$} &\multicolumn{3}{c}{$\sigma^2_{lon}$}\\
                    \cline{4-9}
                    &&&$\mu$ &$\sigma^2$ &$\max$ &$\mu$ &$\sigma^2$ &$\max$\\
                    \midrule
                    A &SS &274 &1.469 &3.604 &22.689 &1.355 &3.328 &21.049\\
                    &FH &56 &0.205 &0.376 &1.087 &0.230 &0.419 &1.455\\
                    \hline			
                    B &SS &1,062 &1.061 &4.343 &57.166 &0.851 &4.065 &59.346\\
                    &FH &77 &0.013 &0.045 &0.223 &0.022 &0.077 &0.341\\
                    \hline	
                    All &SS &1,336 &1.145 &4.204 &57.166 &0.955 &3.929 &59.346\\
                    &FH &133 &0.094 &0.263 &1.087 &0.110 &0.295 &1.455\\
                    \bottomrule
                \end{tabular*}
                \begin{tablenotes}[para]
                    \item[a]{Areas: A - Anaheim Hills, B - North Tustin}
                    \item[b]{Objects: SS - Stop Sign, FH - Fire Hydrant}
                    \item[c]{Displayed values scaled $\times10^{-4}$}
                \end{tablenotes}
            \end{threeparttable}
        \end{minipage}
    \end{center}
\end{table}

\begin{figure}[ht]
	\centering
	\includegraphics[width=\textwidth]{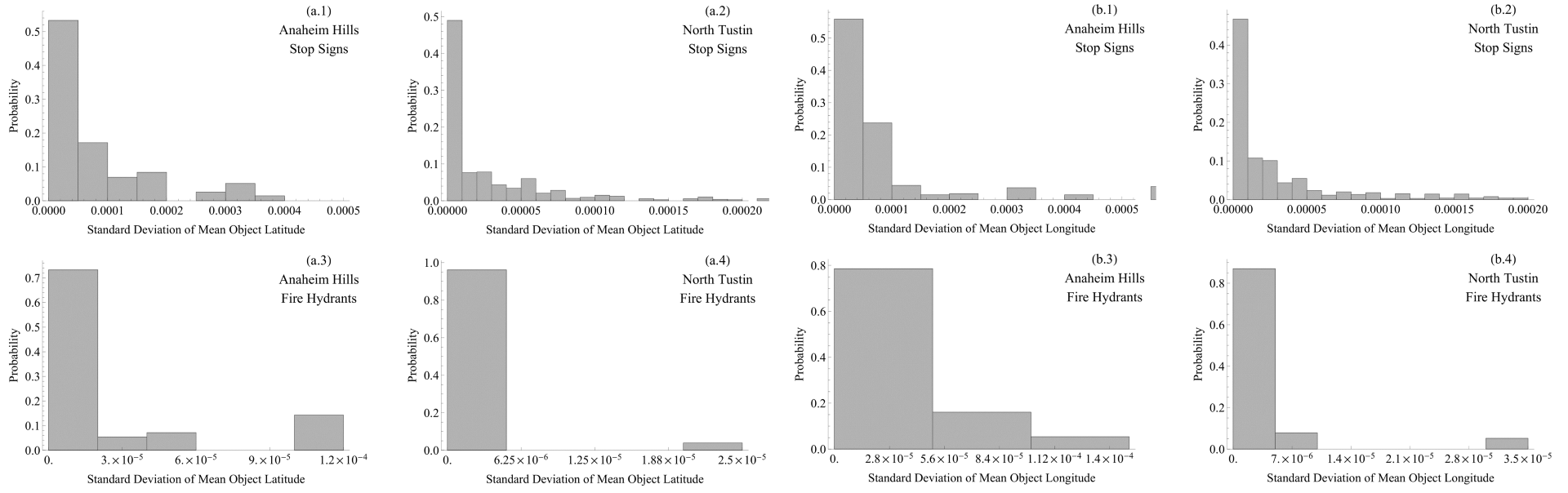}
	\caption{Frequency distribution histograms of the standard deviation in the detected stop sign and fire hydrant object's latitude (subgraph group a), and longitude (subgraph group b). For each subgraph group, the first row represents the stop sign detections, and the second row the fire hydrant detections, while the first column represents the Anaheim Hills sample area, and the second, the North Tustin sample area.}
	\label{fig15}
\end{figure}

As we can see in \figurename{\ref{fig15}}, the standard deviation between multiple detections for each stop sign is relatively low for both latitude and longitude coordinates. More specifically, \tablename{~\ref{table08}} presents the statistics for the mean standard deviation of latitude and longitude estimations for the detected objects (stop signs and fire hydants) in each of the sampled areas. The mean standard deviation of latitude and longitude measurements derive from the variability observed in multiple detections for each unique location of object. The computed statistics (mean value and standard deviation) show the additional variability across all object detections in each area. The mean standard deviation for all stop signs is (0.0001336, 0.0000955) and the mean standard deviation for all fire hydrants is (0.000094, 0.0000110) for the (latitude, longitude) measurements. Thus all measurements present approximate values at the level of $1\times10^{-4}$ range. Such a mean deviation in terms of WGS84 decimal degrees correspond to values around \textpm8 feet at mean point-of-detection to object distance across all detected objects at about 14.33 feet. These results are further bounded by an additional set of confounding factors:
\begin{enumerate}
	\item[a.] The \textit{number of sequential detections} for each object. More detections, further from the object increase variability of the estimates (albeit, potientially improve locational estimates).
	\item[b.] The image's \textit{minimum pixel accuracy ($\theta_{pixel}$)} values as shown in (2).
	\item[c.] The \textit{relative mean distance} between point-of-detection and object. The longer this distance, the largest the within-pixel location resolution error is (thus, pairs of far-object detections are bound to exhibit more variance than pairs of near-object detections).
	\item[d.] The role of \textit{vertical distortion error} exhibited in 360\textdegree~photosphere imagery, as can be seen in \figurename{\ref{fig03}} (albeit our methodology for cardinal photosphere extraction attempts to minimize this error).
\end{enumerate}

\section{Conclusions and Discussion}\label{sec5}

The results and methodology described in this study provide a concise, scientifically accurate, and reliable approach for automatic and spatially explicit object detection using deep learning computer vision algorithms. Our results demonstrate how a systematic data collection methodology that utilizes high-resolution photosphere imagery coupled with both high-end, centimeter-accuracy GPS sensors, and  \textsl{LiDAR} point-cloud datasets can enhance our ability to sense and detect focal objects such as street signs, fire hydrants, and other objects of public interest.

We show how the use of convolutional artificial neural networks for deep learning computer vision tasks can provide reliable and consistent pattern recognition results. Both traditional (pre-trained) models trained on generalized images from the web, as well as the custom computer vision models (for example, Azure and Cloud custom vision models, and the native custom models for TensorFlow/Keras, Matlab, Mathematica, and other applications) achieve acceptable and reliable levels of recognition performance.

The detection methods described in this paper encapsulate some significant benefits and potential for some applications, while presenting challenges and drawbacks for others. For example, the stop-sign detection algorithms can play a critical role in improving efficiency, competency, and accuracy on spatial and geospatial asset inventories. Such applications may reduce cost in labor and time and yield more productive and efficient workflows and eventually public service benefits. The traditional workflows involved for geospatial asset inventories involve laborious and costly field crews with mobile GPS sensors entering data manual-ly. Our showcased methodology allows for accurate, and near real-time detection, positional estimations, and geospatial feature dataset construction.

On the other hand, even for real-time detection algorithms, such models do not have the necessary speed to serve in self-driving and autonomous vehicle settings. As we saw in the example showcased in section IV.B, the maximum detection distance was estimated around 70 feet. Assuming for the sake of this argument that the first detection occurs in real-time, a vehicle must come to a complete stop within the next 70 feet. For a vehicle traveling around 20 miles per hour (29.33 feet per second), this distance translates to a response time a little over 2 seconds. As can be seen in relevant and widely accepted road standards and studies \cite{wu_smart_2009, levinson_towards_2011, savino_evaluation_2013,ilas_electronic_2013}, more likely than not, such response time might not meet the minimum stopping distance (MSD) requirements, even if we ignore human response timing factors. Algorithmic modeling technology for such applications may need to incorporate heuristic, probabilistic and GIS methods in addition to object detection tasks to achieve acceptable levels of performance that meets driving safety standards.

\section*{Acknowledgment}\label{sec6}
The author wishes to thank and acknowledge Kevin Hills, Cameron Smith, and Joe Hunt for their useful suggestions, advice, and insights in both the preparation of this manuscript and the design and implementation of the algorithms used in this study. Earlier versions of this work was presented in a couple of conferences: \cite{alexandridis_deep_2020, smith_how_2019}.


\printbibliography

\end{document}